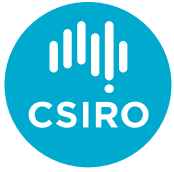

# Automatic Illumination Spectrum Recovery

Nariman Habili, Jeremy Oorloff and Lars Petersson

14 April 2023

# Contents

# 1 Abstract

We develop a deep learning network to estimate the illumination spectrum of hyperspectral images under various lighting conditions. To this end, a dataset, IllumNet, was created. Images were captured using a Specim IQ camera under various illumination conditions, both indoor and outdoor. Outdoor images were captured in sunny, overcast, and shady conditions and at different times of the day. For indoor images, halogen and LED light sources were used, as well as mixed light sources, mainly halogen or LED and fluorescent. The ResNet18 network was employed in this study, but with the 2D kernel changed to a 3D kernel to suit the spectral nature of the data. As well as fitting the actual illumination spectrum well, the predicted illumination spectrum should also be smooth, and this is achieved by the cubic smoothing spline error cost function. Experimental results indicate that the trained model can infer an accurate estimate of the illumination spectrum.

# 2  Introduction

This report discusses the automatic recovery of the illumination spectrum from hyperspectral images captured by a Specim IQ camera. A. To analyse a hyperspectral image, the raw or radiance image must be normalised. The normalised image is known as the reflectance image. To derive the reflectance image, the illumination spectrum of the scene must be known.

Traditionally, the illumination spectrum of a hyperspectral image is recovered by measuring the illumination reflected off a white reference target (or Spectralon) placed in the scene. However, placing a white reference panel in the scene and then estimating the illumination spectrum is a time-consuming endeavour and not always practical. For example, if images are captured from a moving sensor platform (e.g., a vehicle), it is not feasible to use a white reference panel for every image. Moreover, unless the camera allows it (e.g., Specim IQ), illumination estimation will need to be done as part of a post-processing task. To automate the illumination recovery process from an image, a deep learning-based method was developed, and the task was formulated as a regression analysis problem.

The contributions of this research are:

1. The creation of a hyperspectral imaging dataset for illumination spectrum recovery.
2. The use of deep learning to accurately recover the illumination spectrum of hyperspectral images.
3. As part of the deep learning method, ResNet18 was modified to use a 3D kernel suited to modelling spectral data.
4. The adoption of the cubic smoothing spline error function as the loss function.

# 3      Dataset

We used the Specim IQ (Specim Ltd., Oulu, Finland) hyperspectral camera to capture images. The Specim IQ is a handheld hyperspectral camera, which performs hyperspectral data capturing, illumination and reflectance recovery, and visualisation of classification results in one single integrated unit. The sensor uses the push-broom mechanism to capture an image and each image cube is composed of 204 bands with a spatial resolution of 512×512 pixels. The wavelength range of the camera is 400–1000 nm.

The illumination dataset, IllumNet, consists of 1004 images. The images were captured under various lighting conditions and sources, namely sunlight, shadow/overcast, halogen, LED, fluorescent and mixture. The outdoor images were captured at various times of the day to account for changes in sunlight's spectrum. For indoor images, a variety of objects were used, including leaves, fruits, rocks, paper, biscuits, metal, plastic etc, to create complex and diverse scenes. To avoid bias during the training process, the white reference panel was cropped out from all images. Examples of indoor and outdoor images in IllumNet are shown in Figure 1.

It is important to note that our training set is not exhaustive. For example, it does not include images captured in other geographic locations or the use of lighting from different light manufacturers. Our aim in this research was to develop a deep learning network appropriate for illumination spectrum recovery and the network can be retrained with other data to suit the needs of the user.

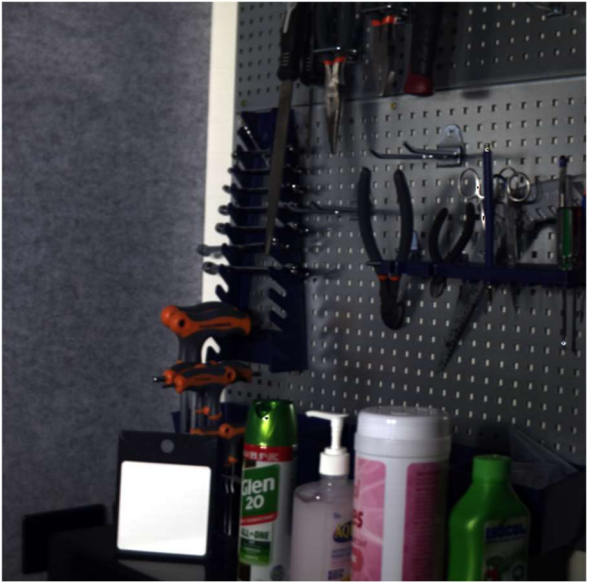
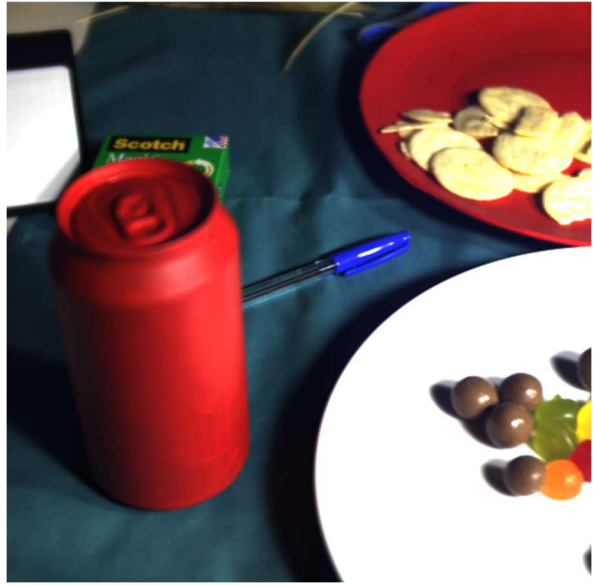
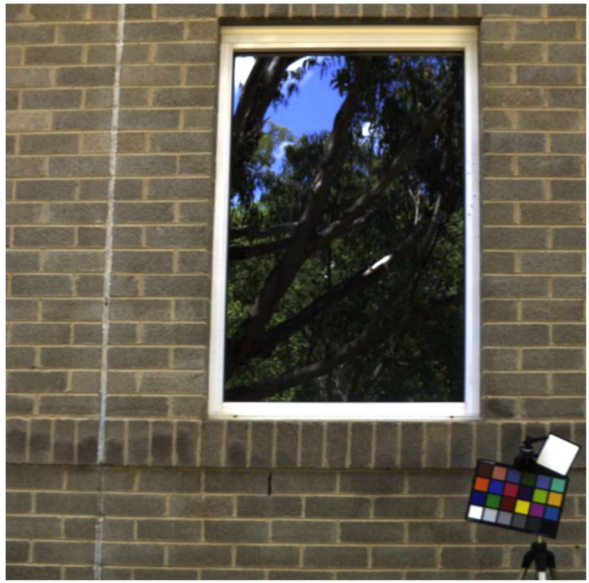
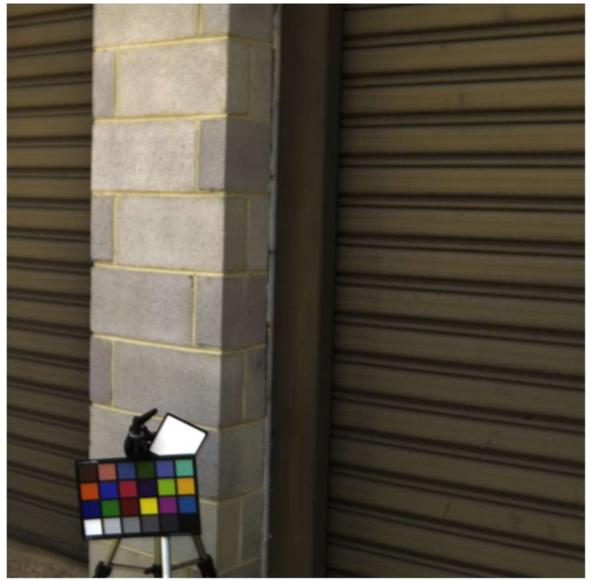

**Figure 1: Examples of indoor and outdoor images captured.**

# 4 Automatic Illumination Spectrum Recovery

## 4.1 Problem Definition

The radiance or raw image captured by a camera is converted to a reflectance image to study the material composition of a scene. The reflectance intensity at pixel $(x, y,)$ for each band can be obtained by:

$$s(x, y, \lambda) = \frac{p(x, y, \lambda) - d(\lambda)}{l(\lambda) - d(\lambda)}$$

where $l(\lambda)$ is the incoming illumination at wavelength $\lambda$, $d(\lambda)$ is the dark reference and $p(x, y, \lambda)$ denotes the radiance intensity. The dark reference represents the baseline signal noise due to the camera's electronics. In the case of the Specim IQ camera, the camera measures this automatically. The most common way to obtain $l(\lambda)$ is to measure the illumination reflected off a white target reference in the scene. The white reference contains material that has a reflectance close to 100% without any spectral features. When the white reference is measured in the same illumination and measurement geometry and distance as the actual material, the signal from the white reference target can be assumed to only contain the signal from the illumination. The white reference target also includes information about the spectral response of the hyperspectral camera, that is, how the camera will affect the measured spectrum.

The goal of automatic illumination recovery is to use deep learning to recover the illumination instead of a white reference target.

## 4.2 Data Augmentation

The illumination recovery dataset consisted of 1004 images, captured with various illumination sources. Of those images, 80% was set aside for training and validation (70% for training and 10% for validation) and 20% for testing. For the training dataset, the white reference targets were cut out to avoid bias during training and $n$ 256×256 sub-images were randomly selected from each image. Each image was then rotated three times (i.e., at 90º, 180º and 360º). To avoid any bias towards either indoor or outdoor images, the training dataset contained approximately an equal number of indoor and outdoor images. The resulting training dataset contained about 40,000 images.

## 4.3 Network Design

In this research, we experimented with several Convolutional Neural Networks (CNN) to recover the illumination spectrum. We experimented with VGG16 [1], ResNet18 [2] and ResNet101 [2]. VGG16 produced large model files, approximately 1 GB in size, which may not be suitable for some portable applications. In some experiments, validation and testing results showed that ResNet18 performs better than ResNet101, probably because it has a shorter network, and a shorter network is more appropriate for the dataset. Therefore, ResNet18 was chosen for further exploration.

The main utility for ResNet is the detection of objects in an RGB image. Recognising that our objective here is not to detect 2D objects, but to co-opt spectral features to recover the illumination of a hyperspectral image, the original ResNet was modified to use 3D convolutions, instead of 2D convolutions. Experimental results demonstrated that significantly better results are obtained by using 3D convolutions. We refer to the modified ResNet network ResNet3D18. The architecture of ResNet3D18 for IllumNet is shown in Table 1. The building blocks of the network, along with the number of nested blocks, are listed in the third column. Note that conv1 has an input channel of 1 and a depth of 51, which is the number of image bands downsampled (by nearest neighbour) by 4. This was done to reduce GPU memory usage. Downsampling is performed by conv31, conv41, and conv51 with a stride of 2. The last layer is a fully connected layer with an output of 204, corresponding to the number of bands in the images.

| Layer name | Output size (channels, depth, W, H) | Building blocks |
|---|---|---|
| conv1 | 64x26x128×128 | 11×7×7, 1 input channel, 51 input depth, 64 output channels, stride 2 |
|  | 64x13x64×64 | Max pool 3×3×3, stride 2 |
| conv2_x | 64x13x64×64 | $\begin{bmatrix} 7 \times 3 \times 3, & 64 \\ 7 \times 3 \times 3, & 64 \end{bmatrix} \times 2$ |
| conv3_x | 128x7x32×32 | $\begin{bmatrix} 7 \times 3 \times 3, & 128 \\ 7 \times 3 \times 3, & 128 \end{bmatrix} \times 2$ |
| conv4_x | 256x4x16×16 | $\begin{bmatrix} 7 \times 3 \times 3, & 256 \\ 7 \times 3 \times 3, & 256 \end{bmatrix} \times 2$ |
| conv5_x | 512x2x8×8 | $\begin{bmatrix} 7 \times 3 \times 3, & 512 \\ 7 \times 3 \times 3, & 512 \end{bmatrix} \times 2$ |

|  |  | Average pool 3D, 204-fc, regression CSSE loss |
|---|---|---|

Table 1: Architecture of ResNet18-3D for IllumNet.

## 4.4 Implementation

The weights were not initialised with any pre-trained network and were trained from scratch. Stochastic gradient descent (SGD) was used with the mini-batch size of 4. Experimental results indicated that low mini-batch values gave better results. The learning rate was set to 0.005, momentum to 0.9 and the models were trained for 100 iterations.

The training and testing workflow of our illumination recovery method is shown in Figure 2.

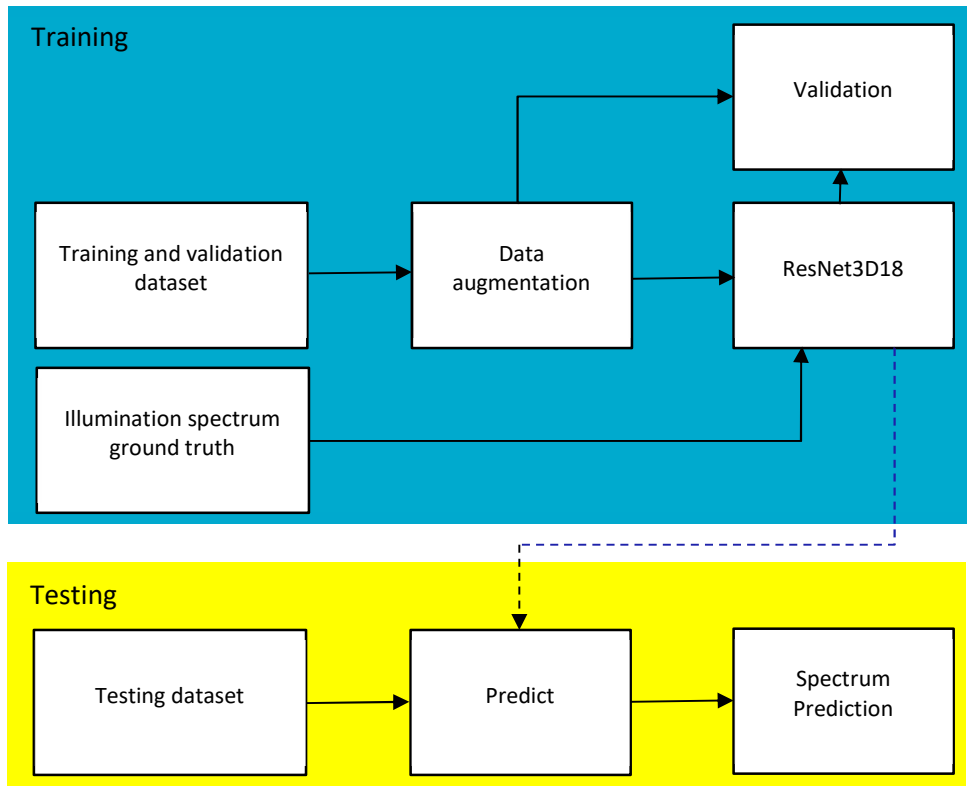

Figure 2: The training and testing workflow for illumination spectrum recovery.

## 4.5 Loss Function

CNNs are trained using an optimization process that employs a loss function to calculate the model error. It is possible to cast the illumination spectrum recovery problem as a regression problem. Example loss functions for regression problems include Mean Squared Error (MSE) and Mean Absolute Error (MAE). The MSE and MAE are computed by:

$$MSE = \frac{1}{N} \sum_{i=0}^{N-1} (y_i - \hat{y}_i)^2$$

and

$$MAE = \frac{1}{N} \sum_{i=0}^{N-1} |y_i - \hat{y}_i|$$

where $N$ is the number of data points, $y_i$ is the spectrum value from the ground truth data and $\hat{y}_i$ is the predicted value for data point $i$. The results showed that both MAE and MSE produced reasonable results, with the predicted spectrum following the shape of the ground truth spectrum well. However, MSE and MAE do not take into consideration the "smoothness" of the spectrum curve and produce rough curves that could result in poor reflectance images.

To obtain a predicted spectrum that is smooth and, at the same time, fits the ground truth spectrum well, we employ the cubic smoothing spline function [3] [4]. Smoothing splines are function estimates, $\hat{f}(x)$, obtained from a set of noisy observations $y_i$ of the target $f(x_i)$ to balance a measure of goodness of fit of $\hat{f}(x)$ to $y_i$ with a derivative based measure of the smoothness of $\hat{f}(x)$. The functions provide a means of smoothing noisy $x_i, y_i$ data.

The cubic smoothing spline estimate $\hat{f}$ of the function $f$ is defined to be the minimiser (over the class of twice differentiable functions) of

$$\sum_{i=0}^{N-1} (y_i - \hat{f}(x_i))^2 + \lambda \int \hat{f}''(x)^2 dx$$

where $\lambda \geq 1$ is a smoothing parameter, controlling the roughness of the function estimate. Note that, $\hat{f}''$ measures the roughness of the function estimate and $\sum_{i=0}^{n-1}(y_i - \hat{f}(x_i))^2$, measures the sum of the squared errors of the function estimate and the observations.

Let us define the forward difference of the predicted values as

$$\Delta \hat{y}_{i+1} = \hat{y}_{i+1} - \hat{y}_i.$$

Using the above equation, we adapt the Cubic Smoothing Spline Error (CSSE) function as a loss function for ResNet3D18 as

$$CSSE = \alpha \frac{1}{N} \sum_{i=0}^{N-1} (y_i - \hat{y}_i)^2 + (1 - \alpha) \frac{1}{K} \sum_{j=0}^{K-1} (\Delta \hat{y}_{i+1} - \Delta \hat{y}_i)$$

where $0 \leq \alpha \leq 1$. Note that $\frac{1}{N} \sum_{i=0}^{N-1}(y_i - \hat{y}_i)^2$ is the MSE. As $\alpha \to 0$, the roughness penalty becomes paramount, and conversely, as $\alpha \to 1$, CSSE approaches the MSE.

The value of $\alpha$ is chosen such that the predicted spectrum is not noisy, and at the same time, is not oversmoothed. It is undesirable that the smoothness of the predicted spectrum is less than the smoothness of the actual spectrum because significant absorption bands, that are usually spikey, might become attenuated.

# 5      Experimental Results and Discussions

Figure 3 shows the error for MSE, roughness and CSSE for various $\alpha$ values on validation data as well as the training error. An interesting observation from the plots is that when $\alpha = 0.6$ and $\alpha = 0.8$, the roughness converges rapidly. When $\alpha = 1.0$, the roughness error does still converge even though we are not minimising for roughness here. Understandably, the roughness values for $\alpha = 1.0$ are always higher.

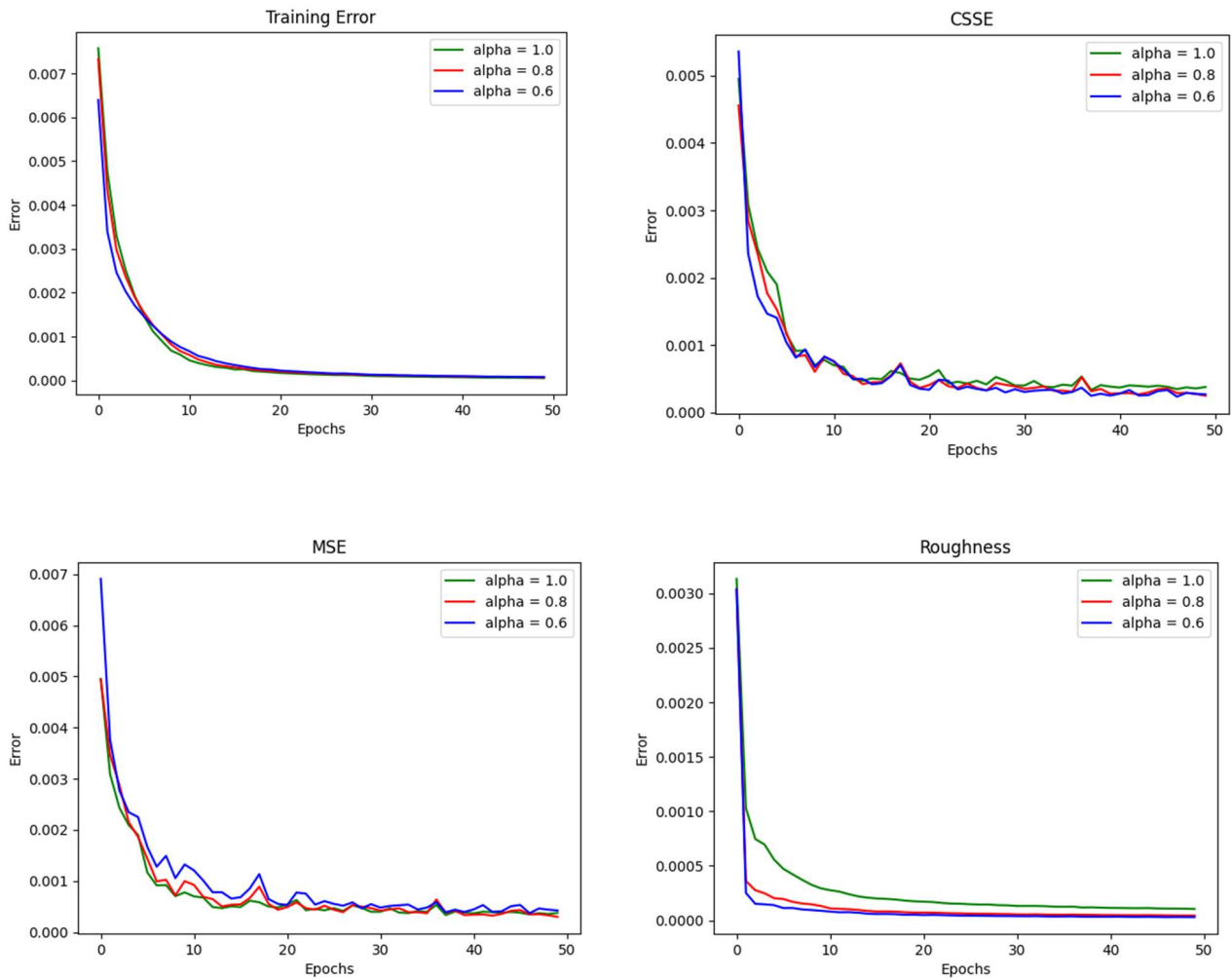

**Figure 3: Training errors on the IllumNet dataset. CSSE, MSE and roughness errors are for the validation dataset.**

Table 2 shows the results for the test data for various metrics and $\alpha$ after 50 epochs. The test set contains 398 full sized images, and this includes both indoor and outdoor images. Interestingly, the lowest MSE is obtained when $\alpha = 0.8$. The lowest roughness is obtained when $\alpha = 0.6$, and this leads to the lowest CSSE. However, using the lowest CSSE to select the best $\alpha$ is not a good idea since a low roughness value might lead to over smoothness of the predicted illumination spectrum. Over smoothing the predicted illumination spectrum may result in eliminating significant absorption bands in the spectrum. The actual average roughness of the test dataset is 0.0000585 and the closest predicted roughness value to this is when $\alpha = 0.8$. The roughness value of $\alpha = 1.0$ is significantly higher, suggesting that a higher roughness value also leads to a higher MSE. The $\alpha$ can be fine-tuned further by training for values $0.6 < \alpha < 1.0$. However, for the rest of this report we will show results for $\alpha = 0.8$

| Metric / α | 0.6 | 0.8 | 1.0 |
|---|---|---|---|
| MAE | 0.0349486 | 0.0343797 | 0.0359198 |
| MSE | 0.0041325 | 0.0039423 | 0.0044582 |
| Roughness | 0.0000346 | 0.0000545 | 0.0001334 |
| CSSE | 0.0024933 | 0.0031648 | 0.0044581 |

Table 2: Results for the test dataset for various metrics and smoothing parameter.

Figure 4 to Figure 9 depict the actual and predicted illumination spectrums for images captured indoor and outdoor under various illumination conditions. In most cases, the predicted illumination spectrum is a close match to that of the actual illumination spectrum. In Figure 6, even though the shape of the two spectra are very similar, their magnitudes are different. This is probably caused by the non-uniformity of the lighting; some regions of the scene are darker than others. The actual illumination spectrum is the illumination spectrum from the white target, whereas the predicted illumination spectrum could be the average illumination spectrum of the scene. The only way of measuring that is by using multiple white targets in the scene.

Figure 9 shows the result of mixing different lighting sources. The image was captured in a room with fluorescent lights and a halogen light source directed at the scene. The spectrum of fluorescent light is spikey due to the use of phosphors in the bulb to attenuate the UV light emitted by the mercury vapour. The actual illumination spectrum is a combination of halogen and fluorescent spectra. The shape of the predicted spectrum is very similar to that of the actual spectrum, but with a different magnitude, and is slightly rougher. This shows that the proposed illumination recovery method can predict a reasonably accurate illumination spectrum even under challenging lighting conditions.

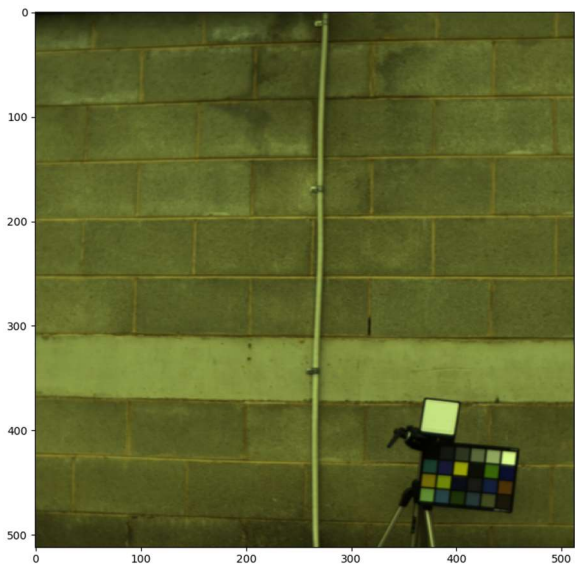
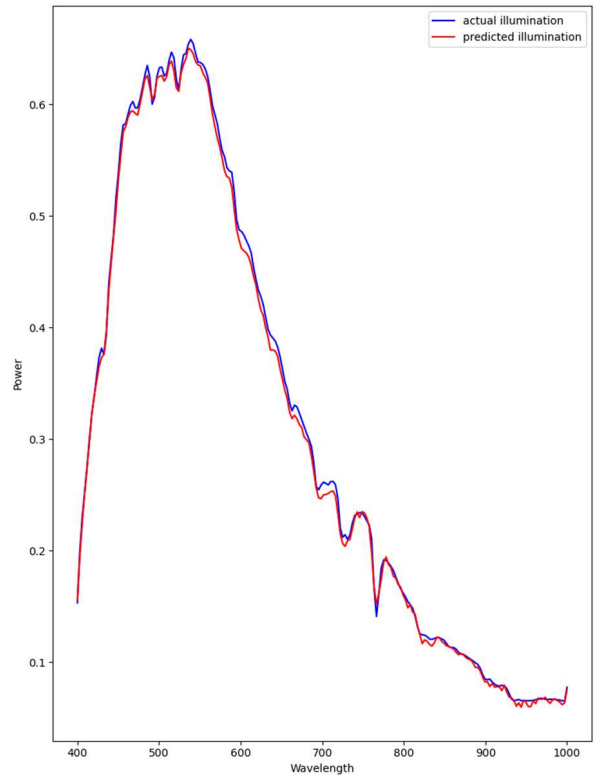

Figure 4: Outdoor and overcast.

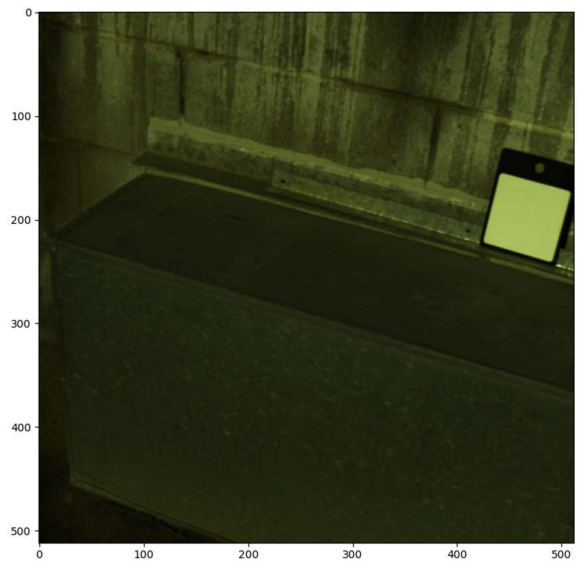
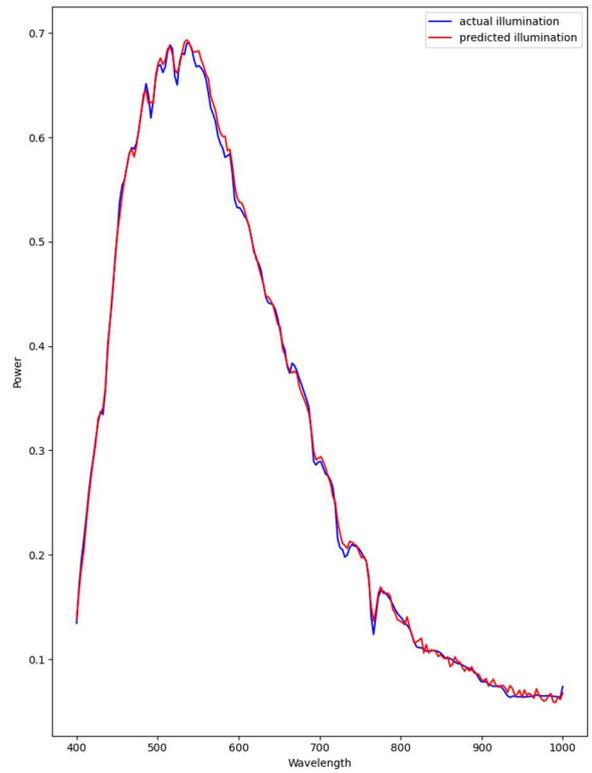

Figure 5: Outdoor and sunny.

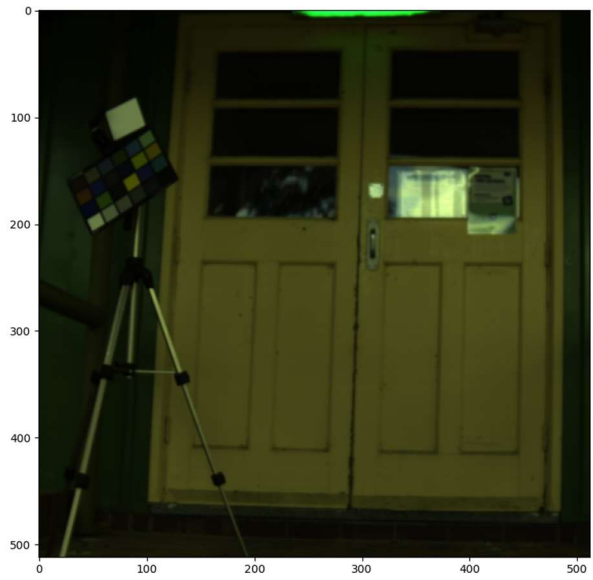 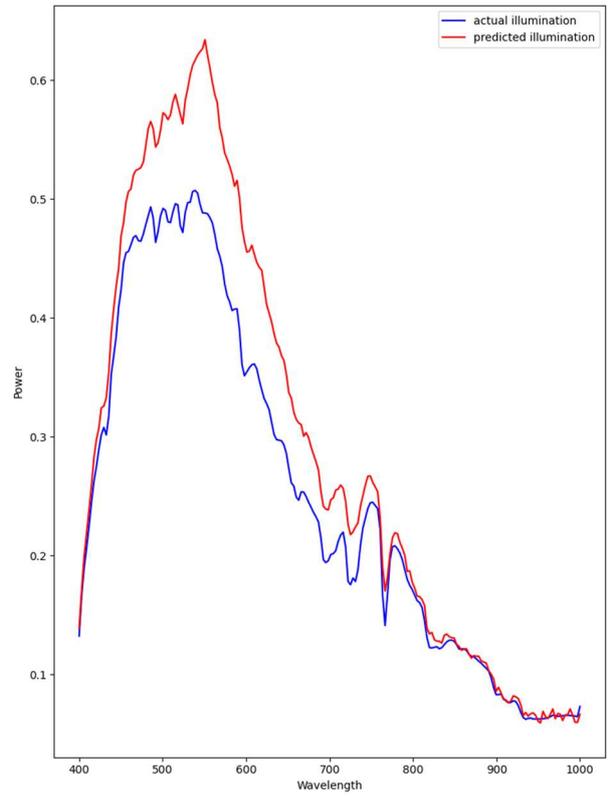

**Figure 6: Outdoor and overcast.**

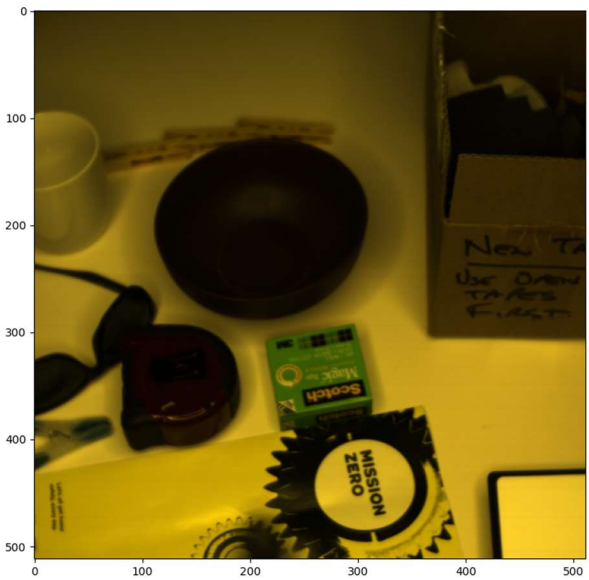 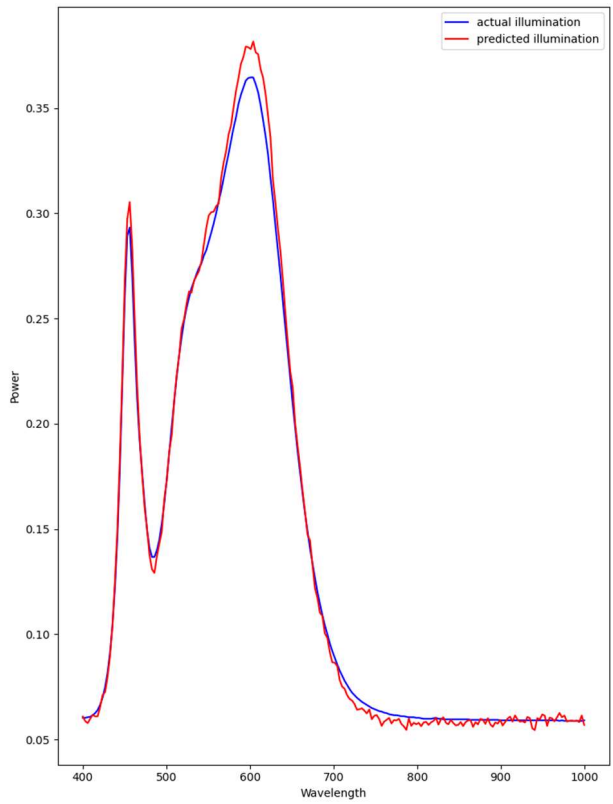

**Figure 7: Indoor and LED.**

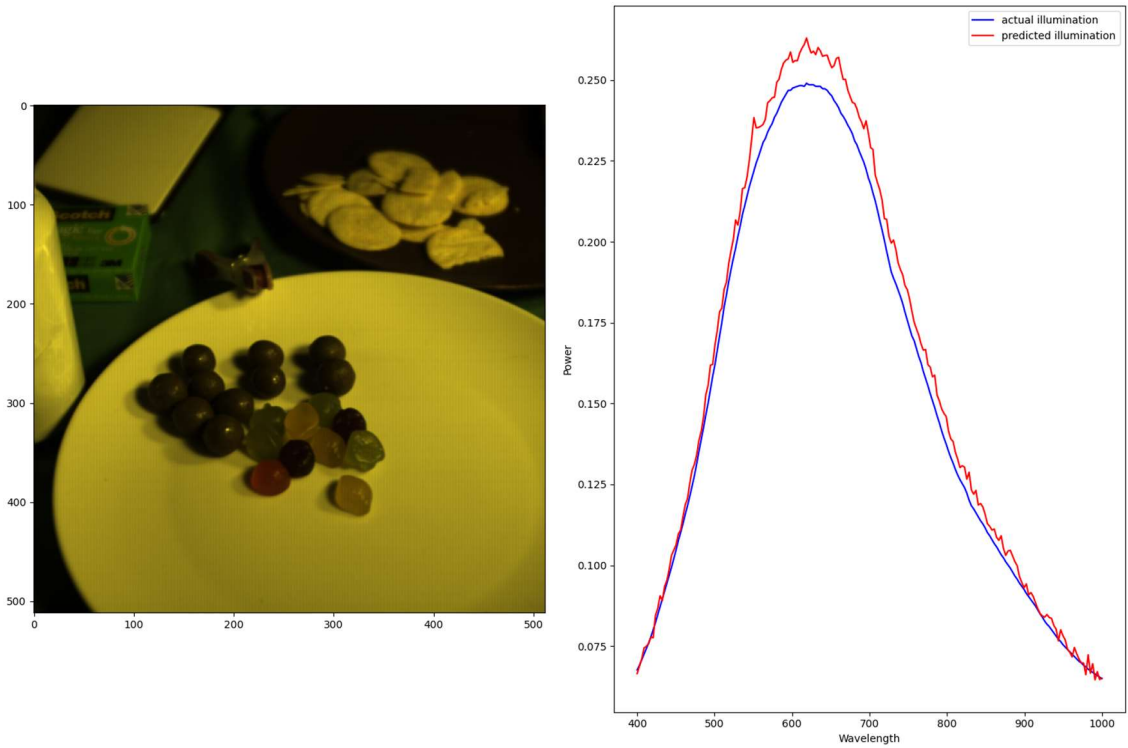

Figure 8: Indoor and halogen.

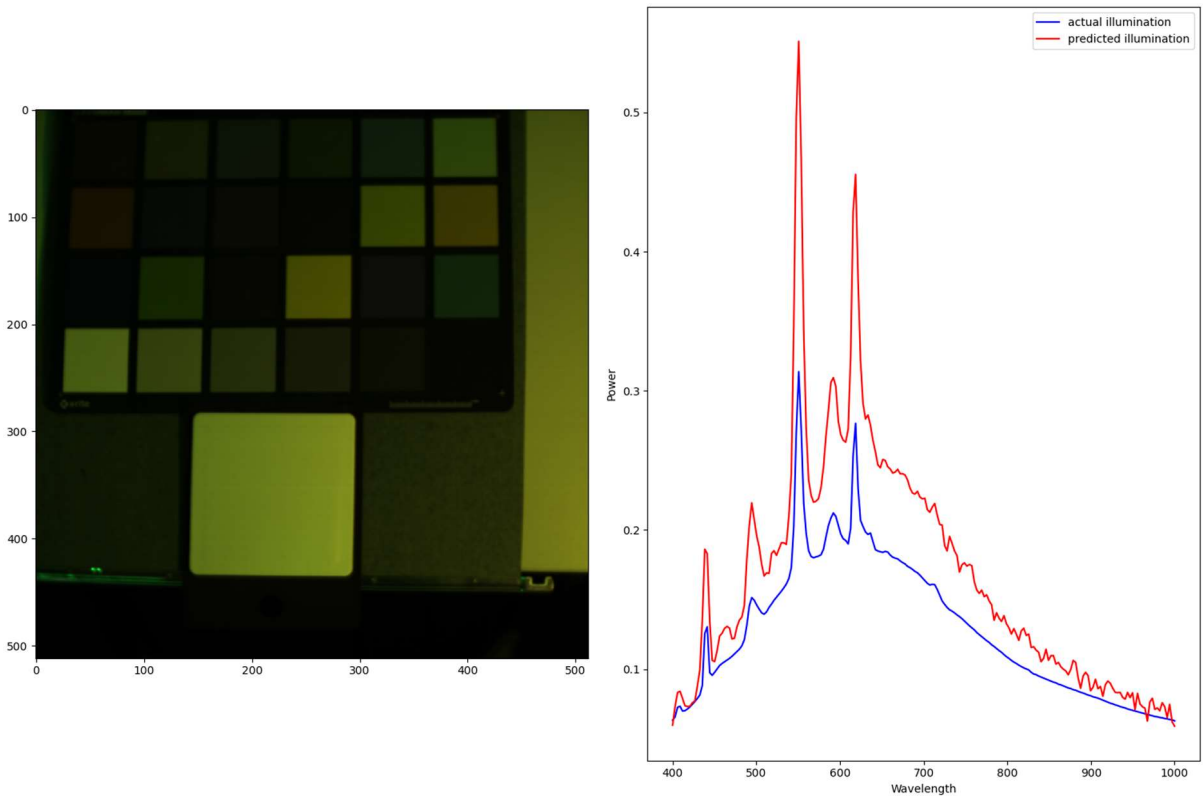

Figure 9: Indoor and halogen and fluorescent mixture.

# 6    Conclusion

The goal of this study was to research and develop AI methods to recover the illumination spectrums of hyperspectral images captured by the Specim IQ camera. A dataset called IllumNet, was created. The dataset contains 1004 images captured both indoor and outdoor, under various lighting sources. The task of illumination recovery was formulated as a regression analysis problem and a deep learning network, based on ResNet18, was developed. ResNet18 was modified to use 3D kernels that better suit the 3D nature of spectral data. A cubic smoothing spline error function was used as the loss function in our deep learning framework. This enables the control of the fit and roughness of the predicted spectrum. Experimental results indicate that the developed deep learning method can recover the illumination spectrum of images.